\documentclass{article}

\usepackage{PRIMEarxiv}

\usepackage[utf8]{inputenc} 
\usepackage[T1]{fontenc}    
\usepackage{hyperref}       
\usepackage{url}            
\usepackage{booktabs}       
\usepackage{amsfonts}       
\usepackage{nicefrac}       
\usepackage{microtype}      
\usepackage{lipsum}
\usepackage{xcolor}
\newcommand{\eqref}[1]{\hyperref[#1]{\textcolor{blue}{Eq.}\textup{(\ref{#1})}}}
\newcommand{\tableref}[1]{\hyperref[#1]{\textcolor{blue}{Table }\textup{(\ref{#1})}}}
\newcommand{\figref}[1]{\hyperref[#1]{\textcolor{blue}{Fig.}\textup{(\ref{#1})}}}

\usepackage{fancyhdr}       
\usepackage{graphicx}       
\graphicspath{{media/}}     

\title{Two-stage Rainfall-Forecasting Diffusion Model
\thanks{\textit{\underline{corresponding author}}: 
\textbf{ChaoRong Li . email: lichaorong88@163.com}} 
}

\author{
  XuDong Ling \\
  Faculty of Artificial Intelligence and Big Data \\
  Chongqing University of Technology ,Yibin University \\
  Yibin 644000,China\\
  \texttt{clearlyzero@stu.cqut.edu.cn} \\
   \And
   ChaoRong Li *\\
   Faculty of Artificial Intelligence and Big Data \\
  Yibin  University\\
  Yibin 644000,China\\
  \texttt{lichaorong88@163.com} \\
 \AND
 FengQing Qin \\
 Faculty of Artificial Intelligence and Big Data \\
 Yibin  University\\
 Yibin 644000,China\\
 \texttt{qinfengqing@163.com} \\
\And
LiHong Zhu \\
Faculty of Artificial Intelligence and Big Data \\
Yibin  University\\
Yibin 644000,China\\
\texttt{zhulihong2022@student.usm.my} \\
\And
Yuanyuan Huang \\
Chengdu University of Information Technology \\
Chengdu 610225,China\\
\texttt{hy@cuit.edu.cn} \\
}

\begin{document}
\maketitle

\begin{abstract}
  Deep neural networks have made great achievements in rainfall prediction.However, the current forecasting methods have certain limitations, such as with blurry generated images and incorrect spatial positions. To overcome these challenges, we propose a Two-stage Rainfall-Forecasting Diffusion  Model (TRDM) aimed at improving the accuracy of long-term rainfall forecasts and addressing the imbalance in performance between temporal and spatial modeling. TRDM is a two-stage method for rainfall prediction tasks. The task of the first stage is to capture robust temporal information while preserving spatial information under low-resolution conditions. The task of the second stage is to reconstruct the low-resolution images generated in the first stage into high-resolution images. We demonstrate state-of-the-art results on the MRMS and Swedish radar datasets. Our project is open source and available on GitHub at:  \href{https://github.com/clearlyzerolxd/TRDM}{https://github.com/clearlyzerolxd/TRDM}.
\end{abstract}


\section{Introduction}
Short-term precipitation forecasting, also known as rainfall nowcasting, delivers detailed predictions of rainfall and hydrometeors within the next two hours\cite{Kann2018STATEMENTOG}. 
With the enhancement of computing power and the continuous development of deep learning methods, recently studies have also employed various architectures of deep neural networks for nowcasting. Shi et al. introduced Conv-LSTM \cite{shi2015convolutional} and Conv-GRU \cite{shi2017deep} models to enhance rainfall accuracy by effectively capturing spatiotemporal correlations in sequence radar data. Conv-LSTM recursives prediction results in higher computational costs. In comparison, Conv-GRU is structurally simplified, reducing network parameters and computational complexity. Both Conv-GRU and Conv-LSTM face the issue of inaccurate long-term predictions. Trebing  et al. \cite{trebing2021smaat} proposed a small attention UNet architecture called SmaAt-UNet, which can effectively capture crucial spatial information within input sequences. However, there is still room for improvement in terms of long-term prediction performance. These networks tend to predict average rainfall rather than actual rainfall, making it challenging to generate reliable rainfall simulations and limiting their effective application in downstream tasks such as hydrological analysis.

Deep neural networks can learn the distribution of rainfall by incorporating generative adversarial networks(GAN)\cite{goodfellow2014generative},  thus generating more realistic rainfall scenes. The latest technology for generating near-term forecasts involves conditional GANs with regularization terms, namely DGMR\cite{skillful}. From the perspective of model training, this network can produce rainfall predictions that are both realistic and accurate. Although the concept of conditional GANs is relatively straightforward, their adversarial training often comes with high and challenging training costs. For instance, differences in the learning speeds of the generator and discriminator may make model convergence difficult and the training process is prone to pattern collapse issues. Moreover, most cGANs no longer utilize the discriminator after training, leading to a certain degree of resource wastage. 
DGMR sacrifices some spatial modeling capability to gain stronger temporal modeling capacity by employing Convolutional Gated Recurrent Units (Conv-GRU) as the temporal processing module. Diffusion models \cite{ho2020denoising} are currently among the most popular generative models. Compared to GANs, they have the advantages of a more stable training process and more diverse sample generation. Rainfall forecasting and video prediction share significant similarities, as both require models to capture spatiotemporal information in order to achieve accurate predictions. In the training and prediction processing of generating high-resolution long videos, the most challenges is insufficient computing resources. Although methods like VAEs\cite{van2017neural} can be used to compress high-resolution videos into lower dimensions for downstream tasks, such models often have extremely large parameters and thus still require significant computational resources to train. To overcome the issues of these methods, we use two difiiusion models to accomplish the task of rainfall prediction. The first model focuses on capturing and analyzing temporal characteristics to accurately predict the timing of future rainfall at low resolution. In order to enhance the spatial dimensional integrity of the prediction, the second model is performed by reconstructing a high-resolution image from a low-resolution image. By decomposing the rainfall spatiotemporal prediction task into a temporal diffusion model and a spatial diffusion model, we are able to effectively predict rainfall scenarios over the next 80 minutes. This layered modeling strategy allows us to more fully and accurately capture the complexity of rainfall variability.
We summarize the contributions as follows:
1.We present a highly efficient architecture for predicting rainfall.
2.Our contribution includes the introduction of a sophisticated 3D sequence Diffusion Model designed for accurate rainfall sequence prediction.
3.Additionally, we put forth two advanced super-resolution methods tailored for reconstructing rainfall images.
\begin{figure*}[t]
 
	\centering	
	\includegraphics[width = \linewidth]{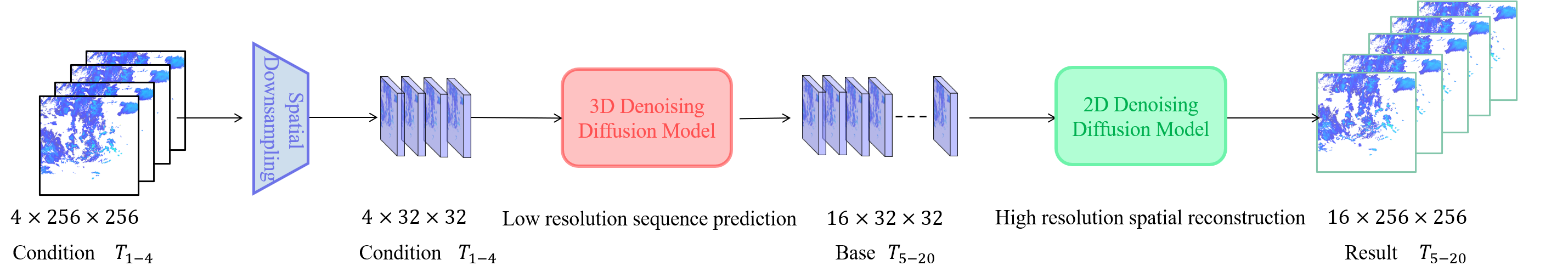}
	\caption{The framework of TRDM. The first part involves a low-resolution diffusion model for prediction, which is comprised of a 3D denoising diffusion model; it uses the input of four conditional frames (20 min) to predict the low-resolution state   ($32 \times 32$) for the next 16 frames(80 min). The second part deals with a super-resolution diffusion model, which is constructed using a 2D denoising diffusion model, it aims to reconstruct low-resolution frames to high-resolution images ($ 256 \times 256 $).
	}
	\label{fig_}
  \end{figure*}
\begin{figure}[h]
	\centering	
	\includegraphics[width = \linewidth]{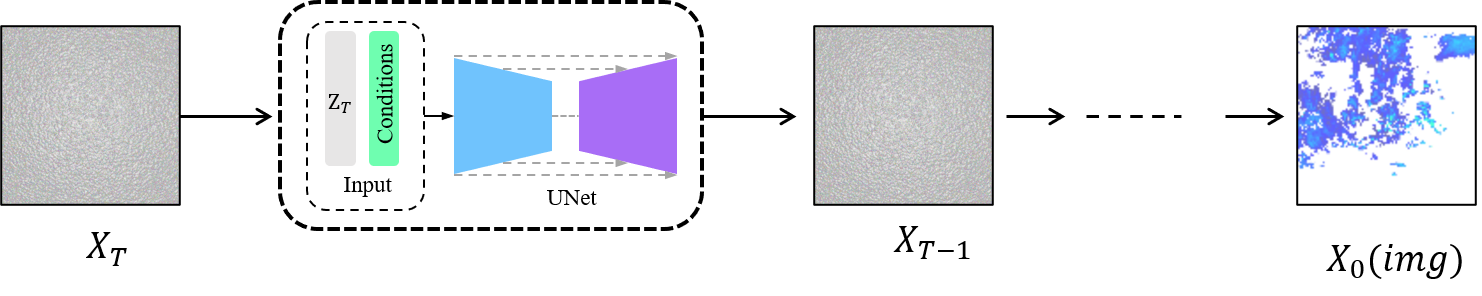}
	\caption{Illustration of SSR inference processing.The diffusion process is performed in pixel space, and the  dimension of $X_i$ are kept consistent with the resolution of the image.}
	\label{fig_diffsuion}
  \end{figure}

  \begin{figure}[h]
	\centering	
	\includegraphics[width = \linewidth]{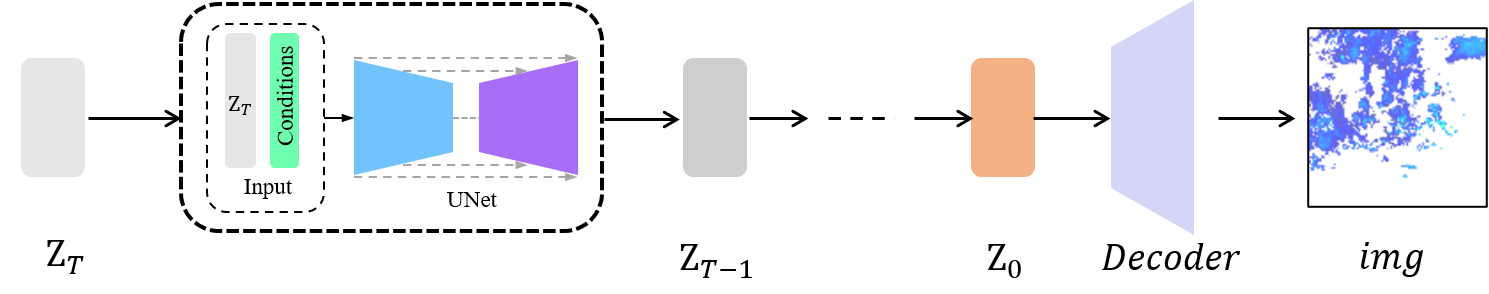}
	\caption{Illustration of LSR inference processing.The diffusion process is performed in the latent space, and the dimension($4\times32\times32$) of $Z_i$ is consistent with the latent variable resolution saved using the encoder.}
	\label{fig_diffsuion_latent}
  \end{figure}

\section{Method}
Our short-term forecasting framework consists of two Stages(As shown in the \figref{fig_}), with the first stage involving the use of a diffusion model to predict the low-resolution rainfall results for the next $M$($X_{5-20}$) frames based on $N$ ($X_{1-4}$) frames of conditions. We refer to this process as the  \textbf{\textit{Prediction Stage}}.
  The goal is to accurately capture the input temporal information and predict future rainfall scenarios at low resolution by applying a sequence conditional diffusion model, while retaining a certain degree of spatial information to provide a robust base for subsequent reconstruction stages.
  \begin{figure*}[!t]
 
	\centering	
	\includegraphics[width = 0.9\linewidth]{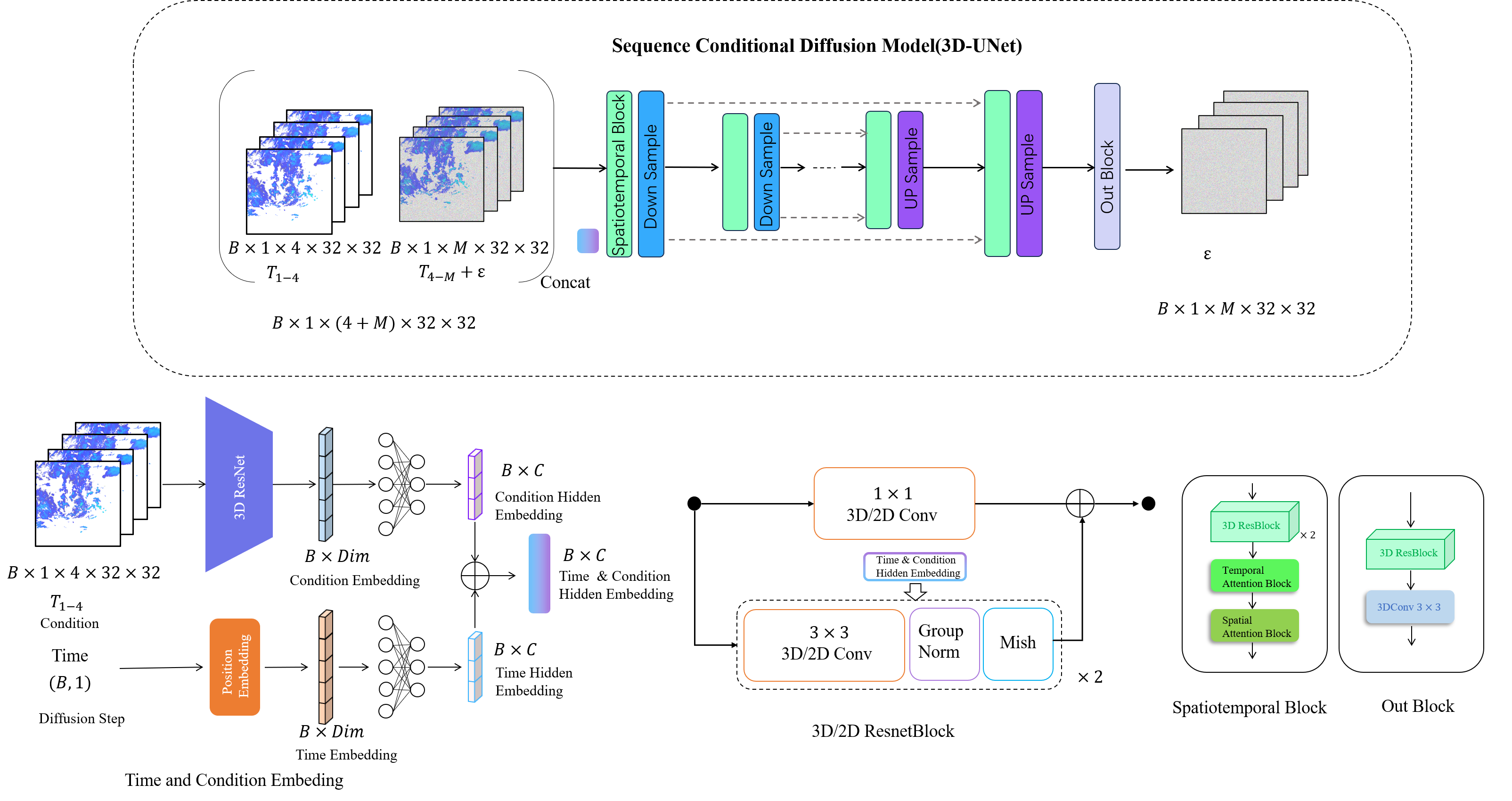}
	\caption{Schematic diagram of training prediction denoising network (3D-Unet)}
	
	\label{Unet_3d}
  \end{figure*}
  In the second stage, we introduced a super-resolution network to reconstruct low-resolution images to high-resolution ones, enhancing image quality and details, enabling us to more accurately analyze the intensity and distribution of future rainfall. We refer to this process as the  \textbf{\textit{Reconstruct Stage}}.
  \subsection{Prediction Stage}
  The prediction Diffusion Model (denoted $\epsilon _{\theta
  }$) is a 3D convolutional neural network(shown in \figref{Unet_3d}) that aims to generate $16$  frames with $32 \times 32$ low-resolution. The inputs of $\epsilon _{\theta}$ involves the sequence of four contextual radar frames $X_{1-4}$ , $X_\epsilon$ obtained by adding noise $\epsilon$ to the 16 frames to be predicted,time information in a diffusion step $t$ time and conditional embedding obtained by the $X_{1-4}$ spatiotemporal features extracted through a 3D-ResNet \cite{hara3dcnns} (denoted as $\tau _{\theta}$).

  $\epsilon _{\theta}$ is trained with $\mathcal{L} _1$ loss. We aim to predict is the  noise, so the loss calculation expression (denoted $\mathcal{L} _{p}$) is defined as \eqref{equ13}:

  \begin{equation}
	\label{equ13}
	\mathcal{L} _{p} = \mathcal{L}_1[\epsilon,\epsilon _{\theta}(X_{1-4},t,X_{\epsilon},\tau_{\theta}(X_{1-4}))]
	 \end{equation}
   \subsection{Reconstruct Stage}
	 In the reconstruct stage, we propose two super-resolution diffusion model. The first model is called spatial super-resolution (SSR), where the training and inference processes are performed directly in high dimensions. The second model is called latent super-resolution (LSR), where the training and inference of the model are performed in low dimensions.
	 \subsubsection{Spatial Super-Resolution(SSR)}
	 The second stage aims to reconstruct the low-resolution image to high resolution. At this stage, we used the diffusion model to build a super-resolution network.
	 The inputs to $\epsilon_{\theta_{ssr}}$  involves the low-resolution image $x_{low}$, time information in a diffusion step $t$ and $x_{\epsilon} $ obtained by adding noise $\epsilon$  to the high resolution. Similar to the loss function used in the prediction network, the training of the super-resolution network also utilizes $L1$ loss, defined as follows \eqref{equ_r}:

   \begin{equation}
    \label{equ_r}
    \mathcal{L} _{SSR} = \mathcal{L}_1[\epsilon,\epsilon _{\theta_{ssr}}(x_{low},t,x_{\epsilon})]
   \end{equation}
   It's important to note that the input low-resolution image ($x_{low}$) is $1\times 32 \times 32$. Before feeding it into the network, it needs to be upscaled to a high-resolution state $1\times H \times W$ using an interpolation algorithm. Afterward, it should be concatenated with the noisy image in the channels and then input into the network together.
   The super-resolution denoising networks employ a 2D-UNet structure, which is similar to the architecture of the prediction network. There are significant differences between them in some key aspects. One of the key differences lies in replacing the 3D-ResBlock with a 2D-ResBlock in this model, and the exclusion of a temporal attention mechanism. This adjustment enhances the network's focus on spatial information.
   \subsubsection{Latent Super-Resolution}
   Inspired by Rombach et al.\cite{rombach2022high}, we proposed a Latent  Super-Resolution method  (LSR) for radar images. LSR is divided into two stages, first by training an AutoEncoder\cite{esser2021taming}(the encoder and decoder are respectively represented as $E$ and $D$) to encode high-resolution images($1\times 256\times 256$) into a latent space(the latent space size is $4\times 32\times 32$). The theory of is almost the same as that of SSR in diffusion processing, Both use low-resolution images as the conditions of the diffusion model($\epsilon _{\theta_{LSR}}$) to guide generate conditions. The difference is that what is generated here is not an pixel space, but a latent space(the schematic diagram is shown in \figref{fig_diffsuion_latent}).  In the train processing, the inputs into $\epsilon _{\theta_{LSR}}$ consists of the low-resolution image $x_{low}$, the implicit encoding of low-resolution image $x_{latent}$ is achieved through the utilization  E to encode low-resolution image, the low-resolution image features(denoted as $x_{emb}$) is extracted using VIT-base \cite{dosovitskiy2020image} to obtain the embedding representation of the image, time information in a diffusion step $t$ and $x_{\epsilon}$ obtained by adding noise $\epsilon$ after encoding at high resolution. The training of the LSR network also utilizes $L1$ loss, defined as follows \eqref{equ_lsr}:
   \begin{equation}
     \label{equ_lsr}
     \mathcal{L} _{LSR} = \mathcal{L}_1[\epsilon,\epsilon _{\theta_{LSR}}(x_{emb},t,x_{latent},x_{low},x_{\epsilon})]
      \end{equation}
      The inference processing of SSR and LSR are consistent with the processes shown in \figref{fig_diffsuion} and \figref{fig_diffsuion_latent} respectively. LSR needs to use $E$ before performing the diffusion step to compress the image from the pixel space to the latent space and then perform the diffusion inference processing, and use $D$ decoding after the diffusion process to restore the hidden space back to the pixel space. Details about these model's structures, as well as training and inference can be found at this \href{https://github.com/clearlyzerolxd/TRDM}{address}.
      \section{Result}
      To evaluate the effectiveness of the approach we proposed, we used radar data from Sweden \footnote{\href{https://www.smhi.se/data/meteorologi/radararkiv}{https://www.smhi.se/data/meteorologi/radararkiv})} and Multi-Radar Multi-Sensitivity \cite{skillful}(MRMS) dataset. And benchmarked the cGANs-based Rainfall (DGMR\cite{skillful}) and the statistical model PySTEPS\cite{pulkkinen2019pysteps}.
      \subsection{Swedish dataset}
      The dataset includes radar data from February 6, 2017, to November 12, 2021. We used data from February 2017 to November 2020 as the training set, and the remaining data was used for validation and testing. 
      \subsubsection{CRPS index analysis}
      We use the CRPS index to analyze the performance(show in \tableref{CRPS}) of the
      TRDM$_{SSR}$,TRDM$_{LSR}$ ,DGMR and  PySTEPS. The results show that the prediction performance of these models is similar in the first 5-20 minutes; however, as time goes by, our proposed tow methods all show the highest accuracy after 20 minutes, with the method that explicitly outperforms. TRDM$_{SSR}$ performs better than the TRDM$_{LSR}$ method. DGMR performs exceptionally well in a short period of time.

      \begin{table}[h]
        \centering
        \caption{CRPS results(Lower is better for CRPS)}
        \label{CRPS}
        
      {\begin{tabular}{llllll}
          \hline
          Model  & 5 min & 20 min & 40 min & 60 min & 80 min\\ \hline
         TRDM$_{SSR}$ &0.1035 &\textbf{0.1155}&\textbf{0.1249}&\textbf{0.1314}&\textbf{ 0.1375}\\
         TRDM$_{LSR}$&0.1211& 0.1244&0.1384&0.1419
        &0.1468
        \\
          PySTEPS
           &0.1062&
           0.1286
           &0.1469
           &0.1588
           &0.1667
          
           \\
        
          DGMR &\textbf{0.0972}& 0.1299&0.14683& 0.1489& 0.1678
          \\

         \hline
        \end{tabular}}
        \end{table}    
      \subsubsection{FSS index analysis}
        The FSS 
        values of different model performances for precipitation (see \tableref{FSS}) exhibit that the TRDM$_{SSR}$ and TRDM$_{LSR}$ achieves the highest average score compared to other models and attains the highest FSS score in the 40 to 80 min intervals. And we observed that the prediction results of the generative model are higher than those of the PySTEPS. TRDM$_{SSR}$ demonstrates strong forecast accuracy over short to medium to long time horizons. The DGMR model is overall stable, while the performance of the PySTEPS model is relatively weak at all time points.

        \begin{table}[h]
          \centering
          \caption{FSS results(Higher is better for FSS)}
          \label{FSS}
          
        {\begin{tabular}{llllll}
            \hline
            Model  & 5 min & 20 min & 40 min & 60 min & 80 min\\ \hline
            TRDM$_{SSR}$&0.9308 &0.9130&\textbf{0.8924}&\textbf{0.8756}&\textbf{0.8603}
             \\
             TRDM$_{LSR}$&0.9212&0.9062&0.8867&0.8704&0.8554
             \\
            PySTEPS
             &0.94699&
             \textbf{ 0.9144}
             &0.8744
             &0.8387
             &0.80710
            
             \\
          
             DGMR&\textbf{0.9414}& 0.9087&0.87605&0.8652&0.8456
            \\

           \hline
          \end{tabular}}
          \end{table} 
      
          \subsubsection{MSE index analysis}
          The performance of various models evaluated using MSE* values (show in \tableref{MSE} ) shows that TRDM$_{SSR}$ method achieves the highest MSE* scores,next is the 
          TRDM$_{LSR}$,DGMR ,and the
          performance of the PySTEPS method on this metric is disappointing.

          \begin{table}[h]
            \centering
            \caption{MSE* results(Normalized the results to 0-1,Lower is better for MSE*)}
            \label{MSE}
            
          {\begin{tabular}{llllll}
              \hline
              Model  & 5 min & 20 min & 40 min & 60 min & 80 min\\ \hline
              TRDM$_{SRM}$ &0.0596&0.07154&\textbf{0.08485}&\textbf{0.09547}&\textbf{0.10524}\\
              TRDM$_{LDM}$&0.06362&0.0737&0.0865&0.0970&0.10666
             
               \\
             PySTEPS
               &0.0925&
               0.10206
               &0.1131
               &0.1219
               &0.1284
              
               \\
            
              DGMR &\textbf{0.044}& \textbf{0.0677}&0.0896&0.0981&0.1095
              \\

             \hline
            \end{tabular}}
            \end{table}

            \subsubsection{CSI index analyze}
            The CSI values of different model performances for precipitation greater than 0.06 mm/h is showed in \tableref{CSI} exhibit. Those models obtained the highest average score model and obtained the highest CSI score in the 20-80 minute interval compared to other models. Among them, TRDM$_{SSR}$ obtained the best results
            \begin{table}[!h]
              \centering
              \caption{CSI results of precipitation above 0.06 mm/h }
              \label{CSI}
              
            {\begin{tabular}{llllll}
                \hline
                Model  & 5 min & 20 min & 40 min & 60 min & 80 min\\ \hline
                $TRDM_{SRM}$ &0.6211 &\textbf{0.5683}&\textbf{0.5121}&\textbf{0.46863}&\textbf{0.42950}
                 \\
                 $TRDM_{LDM}$&0.5913&0.5480
                 &0.4945&0.4514&0.4118

                 \\
                $PySTEPS$
                 &0.6017&
                 0.5400
                 &0.4709
                 &0.4187
                 &0.3772
                
                 \\
              
                 $DGMR$ &\textbf{0.6344}& 0.5048&0.4166&0.3681&0.3345
                \\

               \hline
              \end{tabular}}
              \end{table} 
              \begin{figure*}[!t]
       
                \centering	
                \includegraphics[width =0.8 \linewidth]{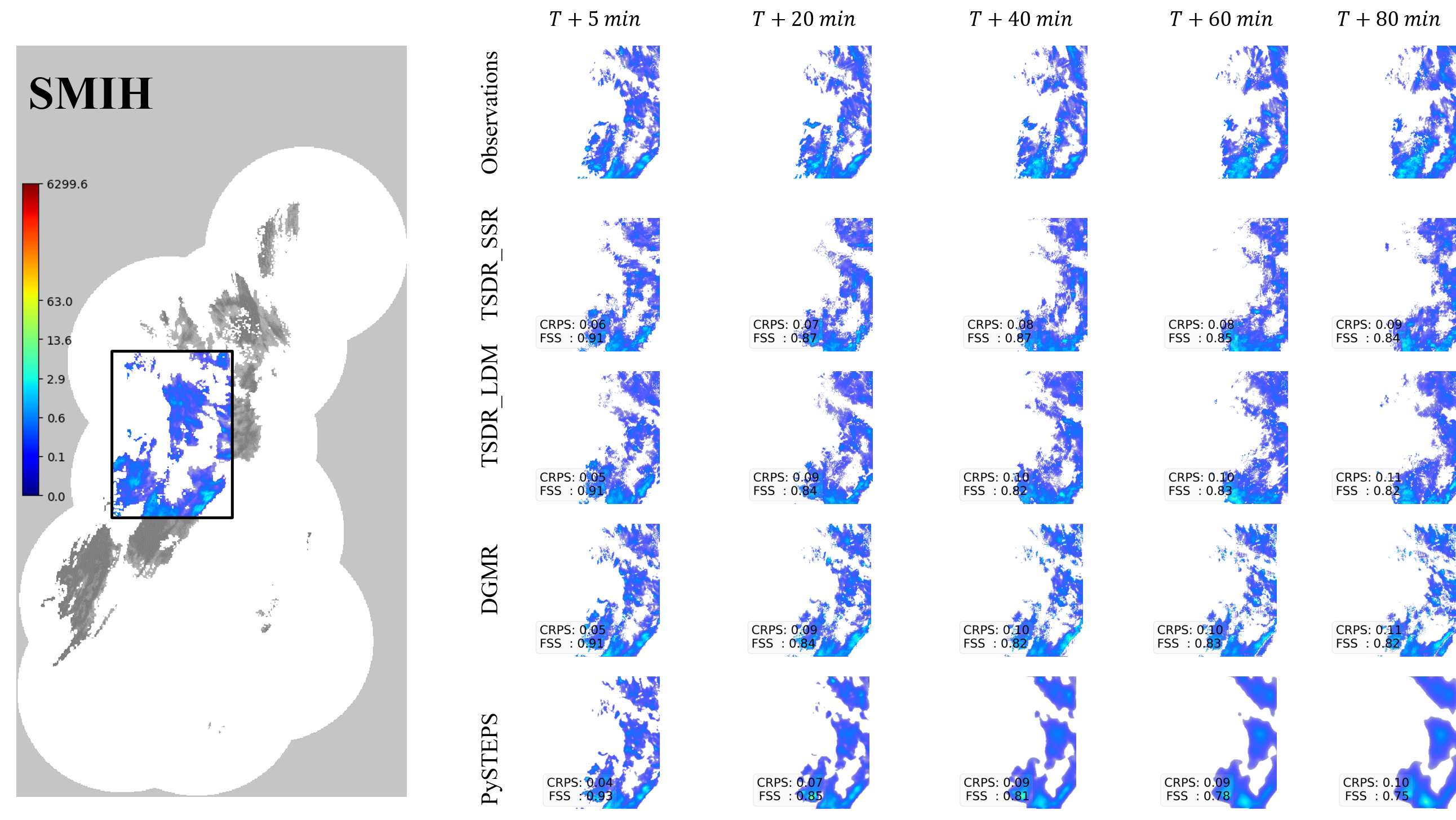}
                \caption{Challenging Precipitation Event Case Study start 2021-01-10 at 03:55 Sweden. Individual predictions at lead times of T+5, T+20,T+40,T+60 and T+80 minutes for different models.The continuous ranked probability score (CRPS) and Field Skill Score (FSS) for each moment displayed in the bottom-left corner.}
                
                \label{result}
                \end{figure*}

                \subsection{MRMS dataset}
                To further substantiate the effectiveness of our proposed method, we have compiled rainfall data for the Multi-Radar Multi-Sensitivity from 2020 to 2022 into a similar dataset. Specifically, the data from 2020 to 2021 is designated for training and validation, while the data from 2022 is reserved for testing.
                \subsubsection{CRPS index analyze}
                We use the CRPS index to analyze the 
                performance(show in \tableref{mrmscrps}) of the
                TRDM$_{SSR}$,TRDM$_{LSR}$, DGMR and  PySTEPS. The performance of these models on this dataset is almost identical to that on the Swedish dataset, with our model achieving the best results after 20 minutes.And the prediction results of the generative model are still better than PySTEPS.
                
                \begin{table}[h]
                  \centering
                  \caption{CRPS results(Lower is better for CRPS)}
                  \label{mrmscrps}
                  
                {\begin{tabular}{llllll}
                    \hline
                    Model  & 5 min & 20 min & 40 min & 60 min & 80 min\\ \hline
                    TRDM$_{SSR}$ &0.1080&0.1342&\textbf{0.1570}&\textbf{0.1711}&\textbf{0.1850}
                     \\
                    TRDM$_{LSR}$ &0.1141&0.1363&0.1573&0.1711&0.1845
                     \\
                    PySTEPS
                     &0.095&
                     0.1354
                     &0.1720
                     &0.1961
                     &0.2122
                    
                     \\
                  
                     DGMR &\textbf{0.086}&\textbf{0.1268}&0.1589& 0.1789&0.1945
                    \\

                   \hline
                  \end{tabular}}
                  \end{table}   
                  
                  \subsubsection{FSS index analyze}
                  The results of FSS indicator calculation is displayed in \tableref{mrmsfss}. At 40-minutes 
                  and beyond, TRDM$_{SSR}$ performs 
                  the best, highlighting the higher accuracy of our proposed model in long-term predictions. In comparison, the PySTEPS model faces some challenges in long-term forecasting. The DGMR model performs well at most time points.

                  \begin{table}[h]
                    \centering
                    \caption{FSS results(Higher is better for FSS)}
                    \label{mrmsfss}
                    
                  {\begin{tabular}{llllll}
                      \hline
                      Model  & 5 min & 20 min & 40 min & 60 min & 80 min\\ \hline
                     TRDM$_{SSR}$ &0.7443&0.6809&0.61407&\textbf{0.5615}&\textbf{0.5087}
                       \\
                       TRDM$_{LSR}$ &0.7556&0.6851&0.6109&0.5541&0.5032
                       \\
                      PySTEPS
                       &0.7820&
                       0.6780
                       &0.5705
                       &0.4959
                       &0.4377
                      
                       \\
                    
                       DGMR &\textbf{0.8498}& \textbf{0.7310}&\textbf{0.6258}& 0.5484&0.4822
                      \\

                     \hline
                    \end{tabular}}
                    \end{table}   
                
\subsubsection{MSE index analyze}
                The result is displayed in \tableref{mrmsmse}. TRDM$_{SSR}$ consistently performs well at 20 and  40minutes, TRDM$_{LSR}$ model  show effectiveness across long time frames. In contrast, PySTEPS underperformed consistently. DGMR excels at 5 and 40 minutes but shows lower performance at other times. Overall, TRDM$_{SSR}$ and TRDM$_{LSR}$model performs consistently well, DGMR excels in shorter time frames, and PySTEPS performs weakly across all time.
                
                    \begin{table}[!h]
                      \centering
                      \caption{MSE results(Lower is better for MSE)}
                      \label{mrmsmse}
                      
                    {\begin{tabular}{llllll}
                        \hline
                        Model  & 5 min & 20 min & 40 min & 60 min & 80 min\\ \hline
                        TRDM$_{SSR}$ &\textbf{0.1454}&\textbf{0.1975}&0.2620&0.2871&0.3249
                         \\
                         TRDM$_{LSR}$ &0.1591&0.2022&\textbf{0.2531}&\textbf{0.2807}&\textbf{0.3157}\\
                        PySTEPS
                         &0.3561&
                         0.3760
                         &0.4174
                         &0.4443
                         &0.4567
                        
                         \\
                      
                         DGMR &0.3604& 0.3687&0.3984&0.4123&0.4212
                        \\

                       \hline
                      \end{tabular}}
                      \end{table}   
  \subsubsection{CSI index analyze}
  The result is showed in \tableref{csimrms}.
  TRDM$_{SSR}$ model excels in medium to long-term forecasts after 40 minutes. PySTEPS performs well overall but shows weaker long-term predictions compared to TRDM$_{SSR}$ . DGMR is strong at 5 and 20 minutes but declines in later time points. In summary, Our proposed two reconstruction methods is best for medium to long-term.

        \begin{table}[!h]
          \centering
          \caption{CSI results(Higher is better for CSI) bove 1 mm/h}
          \label{csimrms}
  
        {\begin{tabular}{llllll}
            \hline
            Model  & 5 min & 20 min & 40 min & 60 min & 80 min\\ \hline
            TRDM$_{SSR}$ &0.4686&0.4082&\textbf{0.3524}&\textbf{0.3101}&\textbf{0.2772}
             \\
             TRDM$_{LSR}$ &0.4691&0.3962&0.3310&0.2865&0.2515
             \\
            PySTEPS
             &0.5291&
             0.4145
             &0.3233
             &0.2668
             &0.2289
            
             \\
          
             DGMR &\textbf{0.5797}& \textbf{0.4407}&0.3529&0.2945&0.2511
            \\

           \hline
          \end{tabular}}
          \end{table}   
  \subsection{Summary}
  We conducted experiments on two datasets and compared them with the most advanced Generative Adversarial network DGMR and the most popular machine learning method-PySTEPS, which fully verified the advancement of our proposed method.
  All methods show the same trend on both datasets, and it can be clearly observed that the generative model outperforms the machine learning method in most indicators and in prediction at most times. Generative models continue to show clear advantages in capturing details and generating realistic images, making better predictions by learning the spatiotemporal distribution information of rainfall data. Within the time frame of the first 20 minutes or so, the prediction performance of our proposed method is similar to that of DGMR. This reflect that for forecasts over shorter time horizons, the two methods have similar effects in capturing temporal correlations. However, over time, our approach gradually showed superiority. This is attributed to the combined advantages of our two-stage diffusion model in temporal and spatial correlations. Among them, the two super-resolution models used in the second stage have little impact on the results, and TRDM$_{SSR}$ is slightly better than TRDM$_{LSR}$.
  \section{Conclusion and Future Work}
  In this letter, the two-stage Diffusion Model we designed takes full account of the temporal and spatial correlations to better handle the task of rainfall sequence generation. In the diffusion model, the model focuses on predicting low-resolution rainfall sequences, focusing on temporal correlations. This allows the model to better capture the evolution of rainfall over time, thereby improving prediction accuracy for future moments.In the second model, super-resolution technology is used to reconstruct  low-resolution rainfall images to high resolution. The focus of this model is on spatial correlation. By learning the spatial structure and details in the image, the reconstructing high-resolution images is closer to the real rainfall field, thereby improving the accuracy of rainfall prediction. In this way, we not only pay attention to the importance of temporal changes, but also fully consider the spatial characteristics of rainfall images, making the generated high-resolution images more detailed and realistic. Our future research will focus on enhancing prediction accuracy over all periods of time and improving inference speed.

\clearpage
\bibliographystyle{unsrt}  
\bibliography{references}

\end{document}